\documentclass{article}

\PassOptionsToPackage{numbers, compress}{natbib}

\usepackage[preprint]{neurips_2026}

\usepackage[utf8]{inputenc} 
\usepackage[T1]{fontenc}    
\usepackage{hyperref}       
\usepackage{url}            
\usepackage{booktabs}       
\usepackage{amsfonts}       
\usepackage{amsmath}
\usepackage{amssymb}
\usepackage{nicefrac}       
\usepackage{microtype}      
\usepackage[table]{xcolor} 
\usepackage{booktabs} 

\usepackage{graphicx}
\usepackage{capt-of}
\usepackage{placeins}
\newcommand{\methodname}{ReScene}
\title{ReScene: Structured Indoor Scene Reconstruction from  Multi-View Captures
}

\author{%
  Haoran Xu$^{1}$ \qquad
  Lechao Zhang$^{1}$ \qquad
  Daoguo Dong$^{2}$ \qquad
  Yan Gao$^{1}$ \qquad
  Xin Tan$^{1,*}$ \\
  $^{1}$School of Computer Science and Technology, East China Normal University \\
  $^{2}$Institute of Trustworthy Embodied AI, Fudan University \\
  \texttt{10235102427@stu.ecnu.edu.cn} \qquad
  \texttt{51275901049@stu.ecnu.edu.cn} \\
  \texttt{dgdong@fudan.edu.cn} \qquad
  \texttt{ygao@sei.ecnu.edu.cn} \qquad
  \texttt{xtan@cs.ecnu.edu.cn} \\
  $^{*}$Corresponding author
}

\usepackage{enumitem}
\begin{document}

\maketitle
\begin{abstract}
Constructing simulation-ready 3D scenes from multi-view captures is a key bottleneck for Embodied Artificial Intelligence, as downstream tasks require object-level structure, explicit inter-object relations, and physical plausibility. Existing approaches either rely on specialized capture hardware, suffer from single-view bias in object reconstruction, or
yield layouts that are geometrically reasonable but physically
inconsistent. We identify that the problem is not single-object
reconstruction but \emph{cross-view relation fusion and physically
plausible scene assembly}. To address this challenge, we present \textbf{ReScene}, a framework that
threads multi-view geometry throughout the pipeline as a
unifying prior. Our method consists of two main components: \textbf{HierView} prioritizes reconstruction views based on semantic
consistency and 3D coverage completeness, replacing the largest-mask
heuristic that conflates image occupancy with object coverage; and \textbf{Relation-Aware Assembly} fuses multi-frame
relation predictions from a vision-language model with geometric and
room-shell priors into a confidence-weighted scene graph, enabling physically consistent scene assembly. ReScene sets a new state of the art across geometry, rendering, and perceptual quality on a set of ScanNet scenes, , achieving a 17\% reduction in Chamfer Distance and 26\% in LPIPS over the strongest prior baseline, while running up to 10$\times$ faster than prior multi-view methods. Based on the reconstructed scenes, we also generate embodied visual question answering dataset, on which fine-tuned Qwen-VL approach the performance of strong closed-source models on several spatial reasoning tasks.
\end{abstract}

\vspace{-10pt}
\section{Introduction}
High-quality, structurally complete indoor simulation scenes~\citep{kolve2022ai2thorinteractive3denvironment,habitat,deitke2022procthorlargescaleembodiedai} are a
cornerstone of  Embodied Artificial Intelligence (EAI) research, underpinning
policy training, task planning, and interactive reasoning. However, authoring such scenes by hand is costly and labor-intensive. Traditional platforms rely on artist-made assets and
manual or procedural layout~\citep{front3d,objaverse,deitke2022procthorlargescaleembodiedai}, which
limits scale and leaves a persistent gap between simulation and real
world. It naturally raises the question: \emph{how can we automatically
construct simulation-ready scenes directly from casually captured
indoor videos?}

Recent efforts toward automated scene construction convert 3D scans into simulatable replicas~\citep{yu2025metascenesautomatedreplicacreation}, compose room-scale scenes from RGB-D input~\citep{litereality}, or leverage vision-language models for 3D relational reasoning~\citep{gpt4scene}. Yet producing object-level, simulation-ready scenes from multi-view captures remains challenging: existing methods are biased toward single-view reconstruction and often sidestep assembly itself~\citep{simrecon,rico,ni2025decompositionalneuralscenereconstruction,yang2025instascenecomplete3dinstance}. We identify the main bottleneck as cross-view relation fusion and
physically plausible assembly, as each viewpoint provides only partial
geometry, leading to inconsistent relations and unstable scene layouts.

\begin{figure}[t]
    \centering
    \includegraphics[width=\linewidth]{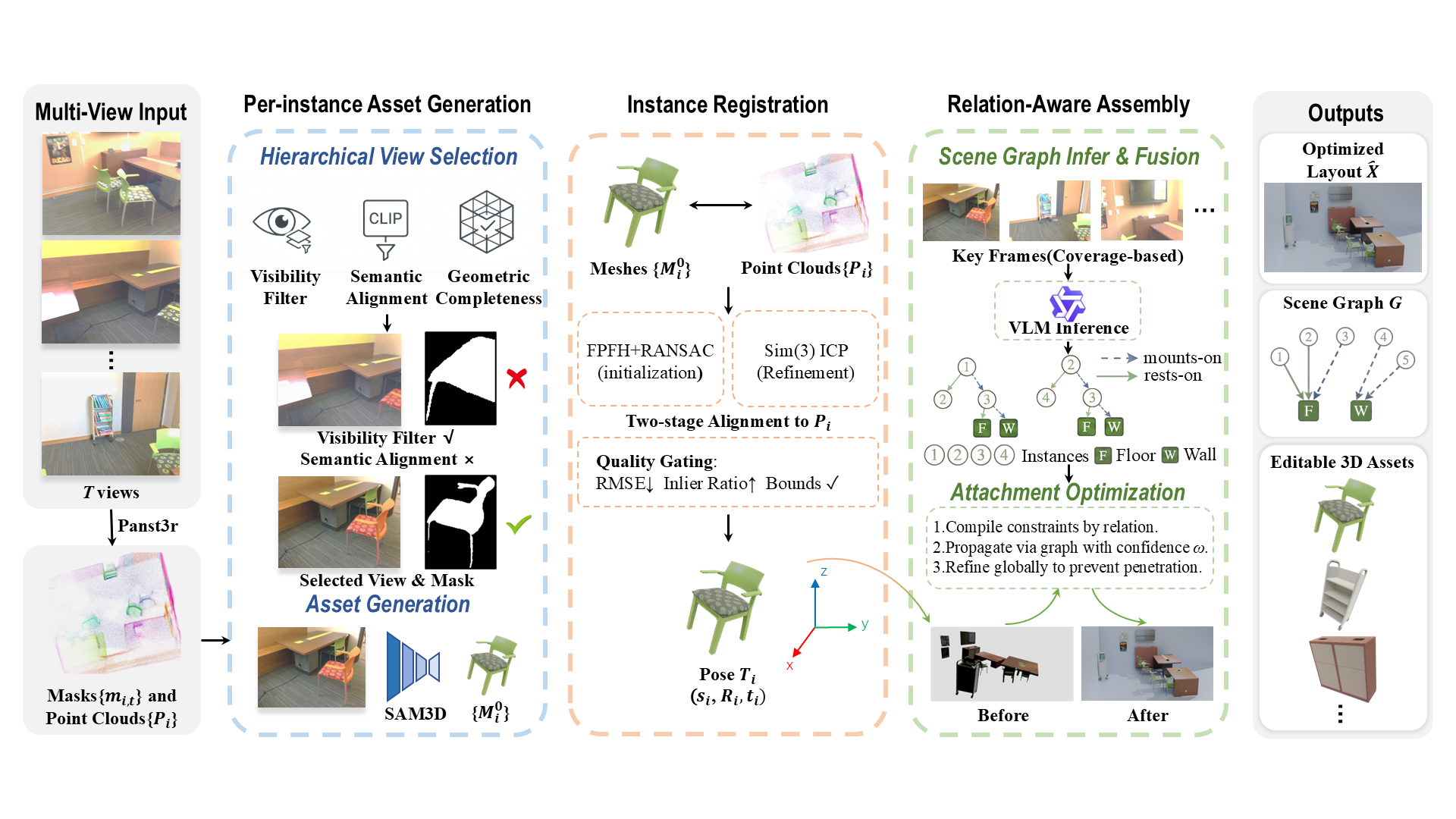}

    \caption{Overview of our framework. Given a casually captured room
    video, our method proceeds through three stages:(1) per-instance asset generation with HierView (from multi-view reconstruction outputs), (2) instance registration, and (3) relation-aware assembly.}
    \label{fig:pipeline}
\end{figure}
Building on this observation, we propose \methodname. As illustrated in
Figure~\ref{fig:pipeline}, given a indoor video or multi-view reconstruction outputs, ReScene proceeds in three stages:
(1) per-instance asset generation with \textbf{HierView}, our
hierarchical view selector; (2) bounded $\mathrm{Sim}(3)$ registration
to align instances in a shared metric frame; and (3)
\textbf{Relation-Aware Assembly}, which fuses cross-frame relations and
performs staged attachment to produce an object-level, physically
plausible scene.

Our method consists of two main components: the first concerns view
selection. The common heuristic of picking the frame with the largest
object mask conflates \emph{how much of the image an object occupies}
with \emph{how much of the object is captured}. A close-up that crops
the chair's legs has a huge mask but reconstructs poorly. HierView
instead checks mask adequacy, verifies semantic consistency across
candidate views via CLIP, and then measures 3D coverage completeness
by projecting the multi-view point cloud back into each candidate. The
result is a view that is both clear and complete. The second concerns
assembly. Relation-Aware Assembly fuses per-frame relation predictions
from a vision-language model with geometric, category, and room-shell
priors into a confidence-weighted scene graph, then compiles the graph
into staged attachment operations spanning floor support, wall
mounting, object support, room containment, and inter-object contact.
Conflicting per-frame predictions are thus resolved into a single
physically consistent layout. Based on the reconstructed scenes, we further construct an agent that exploits the explicit categories, positions, and support relations of our scenes to automatically generate Embodied Visual Question Answering(VQA) data~\citep{das2017embodiedquestionanswering,scanqa,sqa3d} with analytic ground truth about 5,688 question--answer pairs, on which fine-tuned Qwen-VL achieve substantial gains and become competitive with strong closed-source baselines on several spatial reasoning tasks. \noindent\textbf{Our main contributions are as follows :}
\begin{itemize}[leftmargin=1.5em,itemsep=2pt,topsep=2pt]
    \item We propose ReScene, which is an object-level structured 3D scene reconstruction framework
    for real-world indoor video that unifies object reconstruction,
    relation fusion, and physically grounded assembly, with multi-view
    geometry as a consistent prior throughout.
    \item We propose \textbf{HierView}, a hierarchical view selector that picks
    reconstruction views by semantic consistency and 3D coverage
    completeness, addressing a failure mode of prior largest-mask
    heuristics.
    \item We propose \textbf{Relation-Aware Assembly}, a strategy that fuses multi-frame VLM relation predictions with geometric and room-shell priors into a confidence-weighted scene graph, compiled into staged attachment operations for physically stable layouts.
    \item We propose a scene-driven embodied VQA pipeline that uses object categories, poses, and support relations to generate analytic supervision; fine-tuning on this data improves open-source VLMs to match strong closed-source baselines.
\end{itemize}

\section{Related Works}
\vspace{-5pt}
\subsection{Structured 3D Scene Reconstruction}
\vspace{-5pt}
Early scene reconstruction methods~\citep{gaussiansplatting,nerf,monosdf}
model an entire room as a single radiance or signed-distance field. In
contrast, recent work explicitly factorizes scenes into per-instance
components to support downstream interaction
~\citep{gen3dsr,huang2025midimultiinstancediffusionsingle,meng2025scenegensingleimage3dscene,10.1145/3730841,ni2025decompositionalneuralscenereconstruction,yang2025instascenecomplete3dinstance,yu2025metascenesautomatedreplicacreation,simrecon}.  DPRecon~\citep{ni2025decompositionalneuralscenereconstruction} couples
per-object SDFs~\citep{park2019deepsdf} with score distillation~\citep{poole2022dreamfusion} but relies on heavy optimization and is
sensitive to input quality. InstaScene~\citep{yang2025instascenecomplete3dinstance} segments
instances in 3D Gaussian representation and generatively completes
their geometry, but lacks explicit object placement, leading to frequent
physical invalidities. SimRecon~\citep{simrecon} also targets
simulation-oriented reconstruction with scene-graph guidance and active
view selection recently, but produces coarse layouts and becomes unstable in
highly cluttered scenes. The existing methods move toward object-centric representations, they
treat view selection and relation reasoning as largely
decoupled, leading to inconsistent and unstable scene layouts. In
contrast, ReScene unifies view selection, instance-level $\mathrm{Sim}(3)$
registration, and relation-aware assembly under a shared multi-view
geometric prior, jointly enforcing global consistency and physical stability.

\vspace{-8pt}
\subsection{Image- and Video-Based 3D Asset Generation}
\vspace{-6pt}
Object-level 3D generation has advanced rapidly with diffusion models
~\citep{ho2020ddpm,song2020ddim,rombach2022ldm} and large-scale 3D
datasets~\citep{chang2015shapenet,objaverse,objaversexl,fu2021future}. More recent approaches generate assets natively in 3D by training latent
diffusion transformers over geometric representations
~\citep{peebles2023dit,zhang2024claycontrollablelargescalegenerative,li2025triposghighfidelity3dshape,wu2024direct3dscalableimageto3dgeneration,
yang2025hunyuan3d10unifiedframework,xiang2025structured3dlatentsscalable,li2025craftsman3dhighfidelitymeshgeneration,zhao2023michelangeloconditional3dshape},
sometimes aided by normal-map guidance~\citep{ye2025hi3dgenhighfidelity3dgeometry,li2025craftsman3dhighfidelitymeshgeneration}. Several recent frameworks attempt to
jointly reconstruct multiple objects from a single image and infer their
relative arrangement~\citep{gen3dsr,sam3dobjects,huang2025midimultiinstancediffusionsingle,
meng2025scenegensingleimage3dscene,10.1145/3730841}, with SAM3D demonstrating strong occlusion
reasoning~\citep{sam3dobjects}. However, single-image generation is limited by ambiguous scale and pose,
leading to inconsistent layouts even when objects appear plausible.
Frame-wise application to videos does not resolve this issue, as outputs
remain misaligned without a shared geometric reference. Consequently, it
is insufficient for scene-level reconstruction from multi-view captures.

\vspace{-10pt}
\subsection{3D Indoor Scene Datasets and Simulators}
\vspace{-8pt}

Indoor 3D data exhibits a long-standing trade-off between \emph{realistic
geometry} and \emph{explicit compositional structure}. Scanned datasets such
as ScanNet~\citep{scannet}, Matterport3D~\citep{chang2017matterport3d},
ARKitScenes~\citep{baruch2021arkitscenes}, ScanNet++~\citep{scannetpp},
Replica~\citep{straub2019replica}, and SceneNN~\citep{hua2016scenenn}
faithfully capture real environments with semantic annotations, providing limited support for object-level
manipulation and relational reasoning required by
embodied agents~\citep{kim2024openvlaopensourcevisionlanguageactionmodel,shah2022lmnavroboticnavigationlarge}. In contrast, simulator-oriented platforms construct scenes from discrete,
interactable assets. Artist-authored environments such as
BEHAVIOR~\citep{srivastava2021behaviorbenchmarkeverydayhousehold}, Habitat~3.0~\citep{puig2023habitat3},
and the Behavior Vision Suite~\citep{ge2024behaviorvisionsuitecustomizable} provide rich physics
but scale poorly and deviate from real-world statistics, while procedural
or learned generators~\citep{deitke2022procthorlargescaleembodiedai,paschalidou2021atiss,
yang2024holodecklanguageguidedgeneration,tang2024diffuscene} produce large numbers of object-centric
rooms whose geometry and layouts remain synthetic in distribution.
Consequently, existing datasets and simulators rarely offer both realistic
geometry and native compositional structure. ReScene bridges this divide by reconstructing real captures into instance-level assets organized by an explicit scene graph and physically consistent layout.
\vspace{-10pt}
\section{Method}
\vspace{-8pt}
We present \textbf{\methodname}, a framework that turns a multi-view indoor capture into an editable, object-centric scene. We first formalize the problem (Sec.~\ref{sec:formulation}), then organize the pipeline into three stages, all driven by multi-view geometry as a shared prior: (1) per-instance asset generation guided by multi-view priors via \textbf{HierView} (Sec.~\ref{sec:view_selection}); (2) instance-level Sim(3) registration to a shared metric frame (Sec.~\ref{sec:registration}); and (3) \textbf{Relation-Aware Assembly} (Sec.~\ref{sec:assembly}), which infers a multi-frame scene graph and compiles it into a physically consistent layout. We assume calibrated cameras and a coarse set of dominant planar surfaces (floor and vertical walls) estimated from the input scene geometry.

\vspace{-8pt}
\subsection{Problem Formulation}
\label{sec:formulation}
\vspace{-8pt}

Given an input image sequence $\mathcal{I} = \{I_t\}_{t=1}^{T}$ with calibrated cameras, \methodname{} jointly produces three outputs in a shared world frame: per-instance 3D assets and poses $\mathcal{X} = \{(M_i, T_i)\}_{i=1}^{N}$, a global scene graph $\mathcal{G}$ over object instances and structural roots, and an attachment-optimized layout $\hat{\mathcal{X}} = \{(M_i, \hat{T}_i)\}_{i=1}^{N}$ in which the asset poses have been adjusted to satisfy the relations in $\mathcal{G}$.

Each instance $i$ is associated with a reconstructed mesh $M_i$ and a similarity transformation $T_i = (s_i, R_i, t_i) \in \mathrm{Sim}(3)$. The scene graph $\mathcal{G} = \{(c_k, r_k, p_k, \omega_k)\}_{k=1}^{|\mathcal{E}|}$ has nodes corresponding to object instances together with two structural roots, \texttt{floor} and \texttt{wall}; each edge connects a child $c_k$ to a parent $p_k$ via a relation type $r_k \in \{\texttt{rests-on}, \texttt{mounts-on}\}$, with confidence $\omega_k \in [0,1]$. Since each non-root instance has exactly one incoming edge, $\omega_k$ also denotes the confidence of the unique edge into its child. The optimized pose $\hat{T}_i = \mathrm{Compose}(\Delta T_i, T_i)$ composes a per-instance Sim(3) adjustment $\Delta T_i$ with the registered pose $T_i$, where $\Delta T_i$ is chosen to satisfy relational constraints while minimizing pose deviation from $\mathcal{X}$. All stages share a unified instance index taken from the multi-view reconstruction module~\citep{zust2025panst3rmultiviewconsistentpanoptic}; data flow and missing-instance handling are detailed in Appendix~\ref{appendix:pipeline}.

\vspace{-8pt}
\subsection{HierView: Hierarchical View Selection for Single-View Reconstruction}
\label{sec:view_selection}
\vspace{-8pt}

Single-image 3D reconstruction models such as SAM3D-Objects are highly sensitive to input view quality~\citep{sam3dobjects,gen3dsr}. Existing pipelines typically pick the frame with the largest visible mask~\citep{simrecon,rico}, but this heuristic does not target the actual goal of single-view reconstruction: mask area measures how much of the image the object occupies, while reconstruction needs how much of the object the image captures. A close-up of a chair leg yields a large mask but a partial view; a slightly distant front-facing view yields a smaller mask but captures the full object. HierView casts view selection as a three-stage hierarchical filter, with each stage targeting one failure mode. Let $v$ denote a candidate view (a frame index where instance $i$ is visible).
\vspace{-10pt}
\paragraph{Visibility.} As a hard precondition, we discard candidates whose mask pixel count or bounding-box pixel ratio falls below fixed thresholds (Appendix~\ref{appendix:hierview}). Let $\mathcal{V}_i^{(1)}$ denote the admitted candidates.
\vspace{-10pt}
\paragraph{Semantic alignment.} A surviving view may still be misleading if its mask has drifted to a different category, in which case reconstruction will faithfully recover the wrong object. We score semantic alignment with the target category by encoding the masked image crop and the category text with CLIP~\citep{clip}:
\begin{equation}
\mathrm{Sem}(v; i) = \cos\bigl(\phi_{\mathrm{img}}(I_v \odot m_{i,v}),\; \phi_{\mathrm{txt}}(\ell_i)\bigr),
\label{eq:sem}
\end{equation}
where $\ell_i$ is the natural-language category label of instance $i$, and $I_v \odot m_{i,v}$ zeros out pixels outside the mask. We discard views with $\mathrm{Sem}(v; i) < \tau_s$ to obtain $\mathcal{V}_i^{(2)}$.
\vspace{-10pt}
\paragraph{Geometric completeness.} Among the remaining candidates, we rank by how completely the view captures the object's 3D extent:
\begin{equation}
\mathrm{Comp}(v; i) = \frac{\bigl|\{p \in P_i : \pi_v(p) \in m_{i,v}\}\bigr|}{|P_i|},\qquad
t_i^\star = \arg\max_{v \in \mathcal{V}_i^{(2)}} \mathrm{Comp}(v;i),
\label{eq:completeness}
\end{equation}
where $\pi_v$ is the calibrated projection of view $v$. Unlike mask area, $\mathrm{Comp}(v;i)$ directly measures how much of the object's 3D extent is visible from $v$. The three criteria are applied hierarchically rather than as a weighted sum, since they target different kinds of failure that should not be traded off against each other.
\vspace{-8pt}
\subsection{Instance Registration}
\label{sec:registration}
\vspace{-8pt}

For each instance $i$, we align the initial mesh $M_i^0$ to its semantic point cloud $P_i$ by recovering a similarity transformation $T_i(x) = s_i R_i x + t_i$. We initialize via FPFH feature matching~\citep{fpfh} with RANSAC and refine with a Sim(3) ICP solver~\citep{umeyama}. To prevent ICP drift on partial or noisy point clouds, we impose two trust regions on rotation and a bound on scale:
\begin{equation}
d_{\mathrm{SO(3)}}(\delta R_i^{(\ell)}, I) \leq \theta_{\mathrm{step}}, \qquad d_{\mathrm{SO(3)}}(\Delta R_i^{\mathrm{ref}}, I) \leq \theta_{\mathrm{tot}}, \qquad s_i \in [s_{\min}, s_{\max}],
\label{eq:icp}
\end{equation}
where $\delta R_i^{(\ell)}$ is the per-step rotation update, $\Delta R_i^{\mathrm{ref}}$ is the accumulated refinement rotation relative to the FPFH initialization, and $d_{\mathrm{SO(3)}}$ is the geodesic distance on the rotation manifold. A refined alignment is accepted only if it does not increase nearest-neighbor RMSE, does not significantly reduce inlier ratio, and respects the bounds in Eq.~\ref{eq:icp}; otherwise the pre-refinement pose is retained. Correspondence pruning, quality-gating tolerances, and the conservative ICP fallback for solver failures are detailed in Appendix~\ref{appendix:registration}.
\vspace{-8pt}

\subsection{Relation-Aware Assembly}
\label{sec:assembly}
\vspace{-8pt}

The reconstructed and registered instances are by themselves not yet a usable scene: their relations to one another and to the room shell are still implicit~\citep{threedssg,hydra}. Relation-Aware Assembly resolves this in two coupled steps. We first infer a global scene graph $\mathcal{G}$ by fusing multi-frame VLM predictions with geometric and category priors~\citep{gpt4scene}, then compile each edge into a relation-specific transform and propagate the resulting adjustments along the graph topology.

\paragraph{Per-frame inference.} For each key frame, we render the scene with each visible instance overlaid by its index and prompt a vision-language model (Qwen3-VL-32B-Instruct~\citep{bai2025qwen3vltechnicalreport}) to predict, for every visible instance, a tuple of (id, category, relation, parent). The parent is either another visible instance or one of two canonical structural roots, \texttt{floor} or \texttt{wall}; identifying which specific wall an object mounts on is a geometric question deferred to attachment. Outputs are required to be strict JSON and validated for ID coverage, parent and relation validity, and acyclicity. Key frames are selected by a greedy 3D-coverage criterion over voxelized instance point clouds (Appendix~\ref{appendix:keyframe}).
\vspace{-10pt}
\paragraph{Multi-frame fusion.} For each candidate edge $k = (c_k, r_k, p_k)$ into a child instance:
\begin{equation}
S_k = \alpha\, V_k + \beta\, G_k + \gamma\, \Phi_k, \qquad \alpha + \beta + \gamma = 1,
\label{eq:fusion}
\end{equation}
where $V_k \in [0,1]$ is the voting ratio (the fraction of frames in which $c_k$ is visible and the per-frame VLM prediction agrees with $(r_k, p_k)$), $G_k \in [0,1]$ is a geometric prior selected by parent type (floor contact, wall alignment, or parent surface accessibility), and $\Phi_k \in [0,1]$ is a normalized category-conditioned relation prior. For each instance we select the unique incoming edge $k^\star = \arg\max_k S_k$ and set $\omega_{k^\star} = S_{k^\star}$. Low-confidence selections trigger an A/B reranking pass over high-visibility frames. Weights, the construction of $G_k$ and $\Phi_k$, and the reranking procedure are given in Appendix~\ref{appendix:fusion}.
\vspace{-10pt}

\paragraph{Attachment objective.} The scene graph provides discrete, confidence-weighted edges, but the layout it should induce is continuous. We characterize the desired layout by the following objective tendency:
\begin{equation}
\underbrace{\sum_{(c, r, p, \omega) \in \mathcal{E}} \omega \cdot E_r(c, p)}_{\text{relational fitting}}
\;\downarrow,\qquad
\underbrace{\sum_{i \neq j} E_{\mathrm{pen}}(i, j)}_{\text{non-penetration}}
\;\downarrow,\qquad
\underbrace{\sum_i \|\log(\Delta T_i)\|^2}_{\text{adjustment regularization}}
\;\downarrow,
\label{eq:attachment}
\end{equation}
where $E_r(c, p)$ is the relation-specific energy, $E_{\mathrm{pen}}(i, j) \geq 0$ measures pairwise mesh interpenetration, and $\log(\Delta T_i) \in \mathfrak{sim}(3)$ is the Lie-algebra representation of the pose adjustment. The notation $\downarrow$ indicates that each term should be small under our solution; we do not combine the three into a weighted scalar loss, since binding a child to a specific wall plane, preserving lateral order, and resolving collision clusters are non-differentiable in pose space. Edge confidence $\omega$ scales each relational term, so high-confidence edges induce close-to-full constraint satisfaction while low-confidence edges induce only partial adjustments.
We instead use a rule-based dispatched solver: relation-type compilation reduces $\sum \omega \cdot E_r$, topological propagation prevents redundant accumulation of $\|\log(\Delta T_i)\|^2$ along parent--child chains, and a final non-penetration pass reduces $\sum E_{\mathrm{pen}}$.
\vspace{-18pt}
\paragraph{Compilation by relation type.} Each edge maps to a relation-specific energy $E_r(c, p)$ together with the closed-form minimizer of that energy. The compiled transform $\mathrm{Compile}(r)(c, p)$ used in Eq.~\ref{eq:local_update} is exactly this minimizer. We currently support three relation types: \texttt{rests-on}$\to$floor (snap lower contact surface to the floor plane and clip the footprint to the floor polygon), \texttt{mounts-on}$\to$wall (resolve the canonical wall root to the wall plane minimizing coplanarity and outward-normal residuals, then align while preserving lateral order), and \texttt{rests-on}$\to$object (project the contact surface onto a feasible support face of $p$). A separate non-penetration term $E_{\mathrm{pen}}(i, j)$ applies to all non-related object pairs and is resolved by small contact-reverse displacements. The framework is naturally extensible: introducing a new relation type requires only adding a row to the dispatch table. The full specification of each energy term and its closed-form minimizer is given in Table~\ref{tab:dispatch} (Appendix~\ref{app:dispatch}).
\vspace{-10pt}
\paragraph{Propagation along the scene graph.} A naive per-edge update produces downstream inconsistency: when a parent shifts to satisfy its own constraint, its descendants would be left misaligned in the parent's frame. We therefore solve along the topological order of $\mathcal{G}$, from roots to leaves, so that each child reuses its parent's accumulated adjustment.

For each edge $e_k = (c_k, r_k, p_k, \omega_k)$ in topological order, we compute a confidence-weighted local update on the Lie algebra $\mathfrak{sim}(3)$ and compose it with the parent's accumulated update:
\begin{equation}
U_k = \exp\bigl(\omega_k \cdot \log\bigl(\mathrm{Compile}(r_k)(c_k, p_k)\bigr)\bigr), \qquad
\Delta T_{c_k} \leftarrow \mathrm{Compose}(\Delta T_{p_k}, U_k),
\label{eq:local_update}
\end{equation}
with $\Delta T_{p_k} = I$ when $p_k$ is a structural root. The Lie-algebra scaling recovers the identity at $\omega_k = 0$ and the full transform at $\omega_k = 1$, so the operation is well-defined on the Sim(3) manifold. After the relational layer, a global non-penetration pass is applied to all non-related object pairs (last row of Table~\ref{tab:dispatch}).

\begin{figure}[t]
\centering
\includegraphics[width=\linewidth]{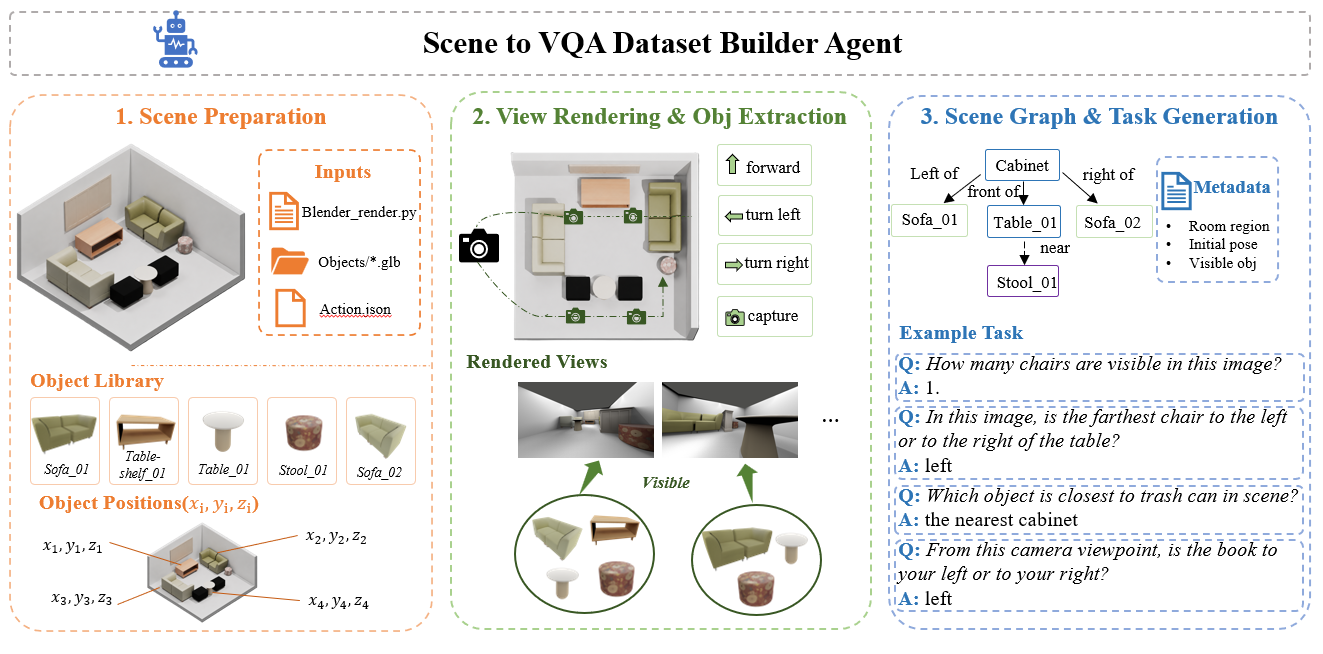}
\caption{
Embodied Visual Question Answering in Assembled Scenes.
Given our structured scene representation, the agent parses object categories, spatial relations, and support/attachment structure to generate Question Answering pairs across multiple reasoning categories.
}
\label{fig:vqa_pipeline}
\end{figure}

\vspace{-15pt}
\subsection{Embodied Visual Question Answering in Assembled Scenes}
\vspace{-8pt}

\label{sec:vqa}
The assembled scene constitutes a complete and physically consistent 3D
environment with explicit object structure, making it suitable for
embodied interaction. To assess whether such scenes can serve as
effective environments for EAI, we design a scene-driven
evaluation pipeline that automatically generates Embodied VQA tasks, as illustrated in
Figure~\ref{fig:vqa_pipeline}.
\vspace{-10pt}
\paragraph{Viewpoint sampling and rendering.}
We implement a headless Blender-based agent that directly loads the
assembled scene and each
scene is composed of instance-level assets with associated position coordinates, along with room floor and wall geometry. The agent operates with a ceiling-centered area light and is initialized
at a random navigable position with a fixed height and random
yaw direction. It executes a predefined sequence of discrete actions
(e.g., forward, turn, look, capture) to traverse the environment and
collect a sequence of egocentric views.
\vspace{-10pt}
\paragraph{View-level visibility extraction.}
At each capture step, the renderer outputs images together with
the set of visible object instances. We determine visibility using ID-based rendering and occlusion-aware
ray casting: an object is considered visible if it exceeds a minimal
pixel threshold in the rendered image, which associates each rendered view with the set of visible object instances.
\vspace{-10pt}
\paragraph{Automatic VQA dataset construction.}
Leveraging the scene graph $\mathcal{G}$, object poses $\mathcal{X}$, and
camera extrinsics, we automatically generate VQA instances for each
captured view. We consider four categories of spatial reasoning tasks:
\emph{object counting}, \emph{distance judgement}, \emph{relation
reasoning}, and \emph{viewpoint estimation}. Ground-truth answers are computed analytically from the structured scene representation,
enabling scalable and fully automatic annotation. To ensure data quality, we perform a lightweight manual inspection on a
subset of generated samples, confirming the correctness of spatial
relations and answer consistency.


\section{Experiments}
\label{sec:experiments}

\subsection{Experimental Setup}
\label{sec:setup}
\paragraph{Dataset.}
We conduct experiments on 30 indoor scenes randomly sampled from the
real-world ScanNet dataset~\citep{scannet}, covering diverse room
types such as bedrooms, living rooms, and offices. Following our
pipeline setting, each scene is processed from multi-view RGB
observations subsampled at one frame every ten (typically 100--200
frames per scene) together with the corresponding camera parameters,
without access to depth, normals, or semantic annotations. Unless
otherwise specified, all experiments are conducted under the same
scene split and preprocessing protocol.

\paragraph{Baselines.}
We compare against multi-view reconstruction methods SimRecon~\citep{simrecon}, RICO~\citep{rico}, and DRAWER~\citep{drawer}, as well as single-view asset-generation baselines Gen3DSR~\citep{gen3dsr} and SAM3D~\citep{sam3dobjects}. All methods use ground-truth camera parameters for absolute-scale alignment, and each baseline is evaluated with its original inputs. InstaScene~\citep{yang2025instascenecomplete3dinstance} is excluded from quantitative comparison due to its closed-source core. Full input specifications are in Appendix~\ref{app:baseline_inputs}.
\vspace{-10pt}
\paragraph{Metrics.}
We evaluate geometry via Chamfer Distance (CD), F@5, and Normal Consistency (NC), rendering via PSNR, SSIM, and LPIPS, and perceptual quality via GPT-5 (see Appendix~\ref{app:gpt_judge} for prompt and scoring). Physical plausibility is measured by object count, out-of-bound (OOB) statistics, and collisions; detailed definitions are in Appendix~\ref{appendix:evaluation}.
\vspace{-8pt}
\subsection{Comparison with State-of-the-Art}
\label{sec:main_results}

\begin{table*}[!t]
\centering
\small
\setlength{\tabcolsep}{4pt}
\renewcommand{\arraystretch}{1.05}
\caption{
Quantitative comparison with state-of-the-art methods on ScanNet, grouped into single-view and multi-view baselines. Metrics include runtime, geometry, rendering, and perceptual scores ($\uparrow$ / $\downarrow$ indicates higher / lower is better). CD: Chamfer Distance ($\times 10^{-2}$). Visual, Complete, and Aesthetic are perceptual scores (1--5) obtained by prompting GPT-5 to assess visual fidelity, scene completeness, and aesthetics. Bold and underline indicate the best and second-best scores, respectively.
}
\label{tab:main_comparison}
\resizebox{\textwidth}{!}{%
\begin{tabular}{l c ccc ccc ccc}
\toprule
& & \multicolumn{3}{c}{Geometry} & \multicolumn{3}{c}{Rendering} & \multicolumn{3}{c}{GPT-5 Score} \\
\cmidrule(lr){3-5} \cmidrule(lr){6-8} \cmidrule(lr){9-11}
Method
& Time (min) $\downarrow$
& CD $\downarrow$ & F@5 $\uparrow$ & NC $\uparrow$
& PSNR $\uparrow$ & SSIM $\uparrow$ & LPIPS $\downarrow$
& Visual $\uparrow$ & Complete $\uparrow$ & Aesthetic $\uparrow$ \\
\midrule
\rowcolor{gray!15}
\multicolumn{11}{l}{\textit{Single-view methods}} \\
Gen3DSR~\citep{gen3dsr}       & \textbf{18.4}  & 20.92 & 30.64 & 51.47 & 13.92 & 0.568 & 0.625 & 1.67 & 1.80 & 1.93 \\
SAM3D~\citep{sam3dobjects}    & \underline{21.6}  & 17.54 & 36.62 & 57.38 & 15.96 & 0.624 & 0.569 & 3.30 & 3.07 & 3.13 \\
\rowcolor{gray!15}
\multicolumn{11}{l}{\textit{Multi-view methods}} \\
SimRecon~\citep{simrecon}    & 28.7  & \underline{10.13} & \underline{63.84} & \underline{76.23} & \underline{18.82}  & 0.697 & \underline{ 0.532} & \underline{4.03} & \underline{3.97} & \underline{3.90} \\
DRAWER~\citep{drawer}        & 375.5 & 11.91 & 63.52 & 70.60 &  16.13  & \underline{0.726} & 0.580 & 2.07 & 2.10 & 2.13 \\
RICO~\citep{rico}            & 262.2 & 11.47 & 62.45 & 74.72 & 18.34  & 0.695 & 0.562 & 3.07 & 2.93 & 2.60 \\
\rowcolor{gray!15}
\multicolumn{11}{l}{\textit{Our Model}} \\
\textbf{Ours}                & 36.3  & \textbf{8.40}  & \textbf{68.99} & \textbf{79.90} & \textbf{19.63} & \textbf{0.770} & \textbf{0.440} & \textbf{4.77} & \textbf{4.80} & \textbf{4.73} \\
\bottomrule
\end{tabular}
}
\end{table*}

\paragraph{Quantitative comparison.}
Table~\ref{tab:main_comparison} shows that our method achieves the best scores across geometry, rendering, and perceptual metrics, while maintaining runtime comparable to the fastest multi-view baseline. Gains are largest on scene-level perceptual scores and LPIPS, indicating that object-centric, relation-aware reconstruction improves both local geometry and global scene completeness. Single-view baselines lag on scene-level metrics, highlighting the importance of multi-view evidence for compositional indoor scenes.

\begin{table*}[!t]
\centering
\small
\setlength{\tabcolsep}{4pt}
\renewcommand{\arraystretch}{1.05}
\caption{
Structural plausibility comparison on ScanNet.
Lower is better for OOB and collision metrics.
}
\label{tab:structural_comparison}
\begin{tabular}{l ccccc}
\toprule
Method
& Obj. Count
& OOB Count $\downarrow$
& OOB Rate $\downarrow$
& Coll. Pairs $\downarrow$
& Coll. Rate $\downarrow$ \\
\midrule
SimRecon~\citep{simrecon}    & \textbf{36.43} & 3.60 & 9.88\%  & 53.27 & 8.25\%  \\
RICO~\citep{rico}            & 10.00 & 2.53 & 25.30\% & 4.50  & 10.00\% \\
\midrule
\textbf{Ours}                & 22.17 & \textbf{0.20} & \textbf{0.90\%} & \textbf{16.10} & \textbf{6.86\%} \\
\bottomrule
\end{tabular}
\end{table*}

\begin{center}
\includegraphics[width=\linewidth]{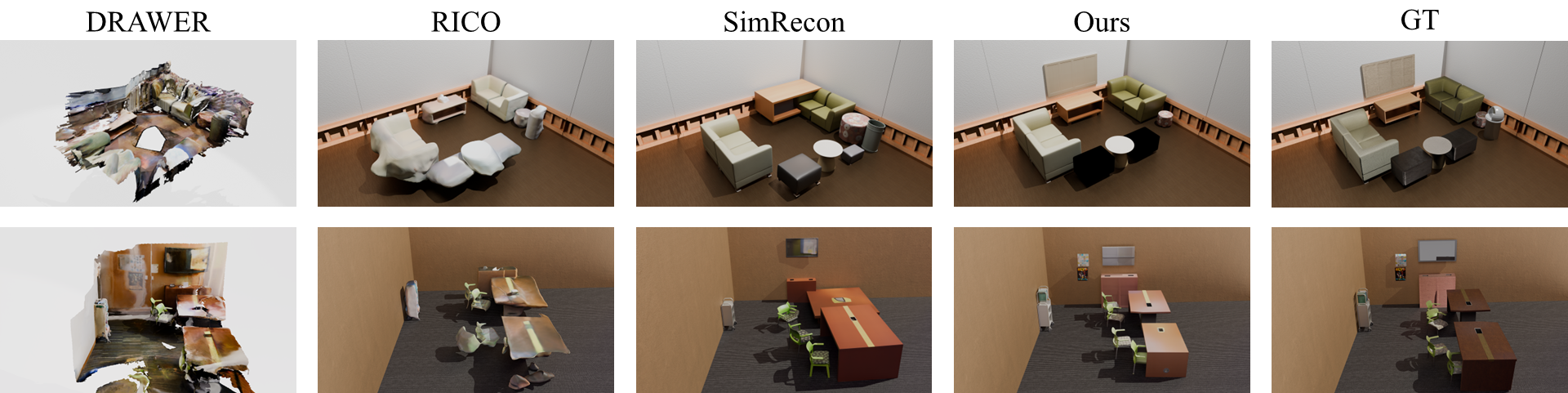}
\captionof{figure}{
Qualitative comparison with baselines on representative ScanNet scenes.
Our method produces cleaner spatial organization, fewer floating objects, and more consistent wall and support contacts, especially in cluttered regions and near room boundaries.
}
\label{fig:qualitative_comparison}
\end{center}

\begin{center}
\includegraphics[width=\linewidth]{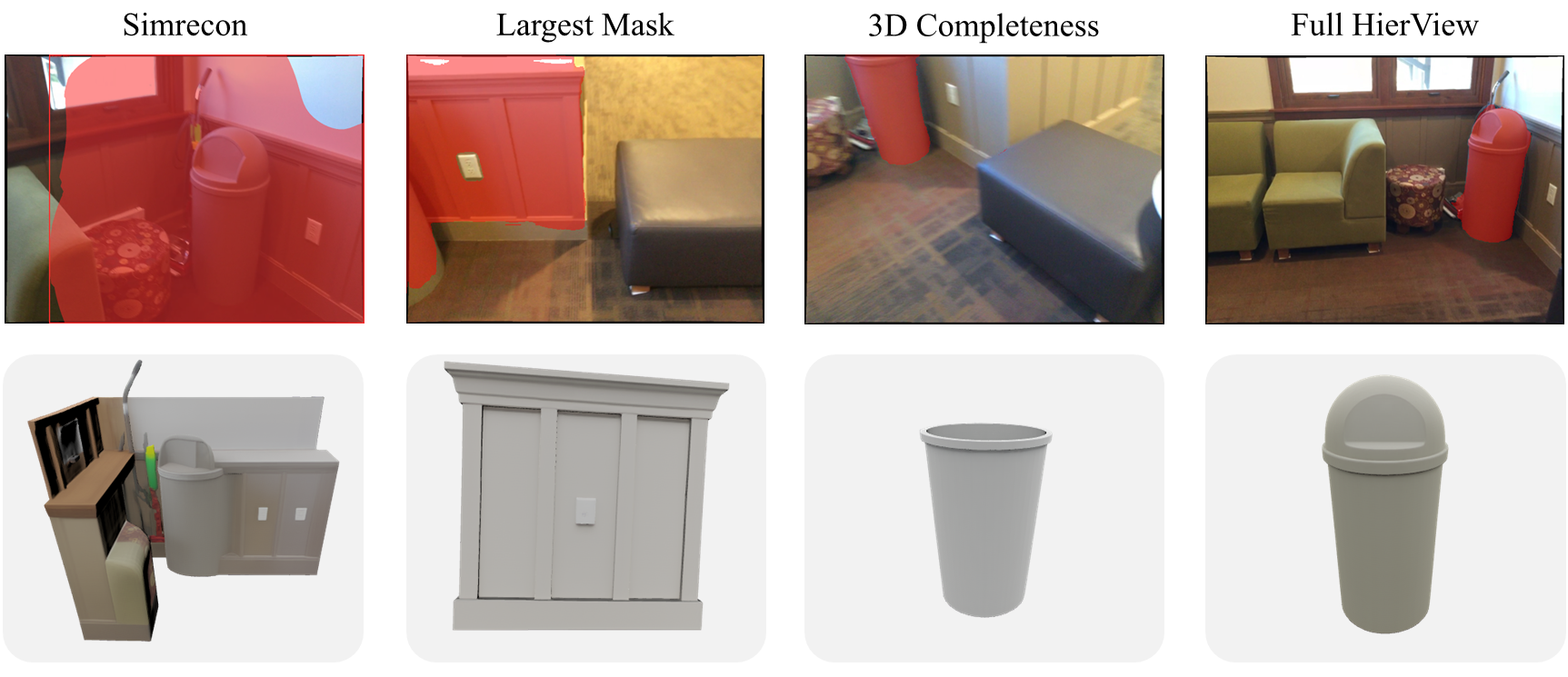}
\captionof{figure}{
Qualitative comparison of view selection and mesh generation. Columns 2--4 show our ablation variants (Largest Mask, 3D Completeness, Full HierView), with column 1 (SimRecon) as reference. Top: instance images/masks; bottom: generated meshes. Without view-quality filtering, raw masks suffer from background leakage and partial-object bias, producing unstable meshes. HierView improves input validity and yields more consistent reconstructions.}
\label{fig:ablation_view_qual}
\end{center}

\paragraph{Qualitative comparison.}
Figure~\ref{fig:qualitative_comparison} highlights failure modes of baselines: SimRecon over-segments and leaves holes on occluded surfaces; RICO merges nearby objects with overly soft surfaces; DRAWER misses small objects and misaligns furniture; single-view methods place plausible assets incoherently. In contrast, our method produces clean, well-separated objects with correct support relations, especially near room boundaries and in cluttered regions.

\paragraph{Structural comparison.}
Table~\ref{tab:structural_comparison} reports structural plausibility metrics for downstream simulation. Our method reduces the out-of-bound (OOB) rate by over an order of magnitude, showing room-shell constraints prevent object drift. While collision rate is modestly lower than SimRecon, absolute collisions drop sharply due to reduced over-segmentation. RICO recovers few objects but still has high OOB rate, showing sparse layouts are not automatically valid. Overall, our method achieves the lowest OOB rate and collision count at moderate object count, confirming that relation-aware assembly produces simulation-ready layouts.

\vspace{10pt}
\subsection{Ablation Study}
\label{sec:ablation}

We ablate the three core modules of our pipeline: view selection, scene graph fusion, and relation-guided attachment. Unless noted, only the target module is changed; all variants share the same scene split and preprocessing. For clarity, \emph{Anchor} denotes the resolved-anchor ratio.

First, we study view selection at the object-level reconstruction stage. Table~\ref{tab:ablation_all}(a) shows three in-framework variants; SimRecon is included as an external reference. 3D Completeness improves CD and NC over Largest Mask by favoring more complete, slightly more distant views, while Full HierView achieves the best overall geometry and rendering quality via semantic filtering. Largest Mask maintains the highest F@5 by selecting close-ups, but at the cost of incomplete objects. Qualitatively (Figure~\ref{fig:ablation_view_qual}), Full HierView produces the most stable object-centric meshes.

Next, we study multi-frame scene graph fusion. An instance is \emph{resolved} if assigned a concrete anchor (its support target) and \emph{unresolved} otherwise. Table~\ref{tab:ablation_all}(b) shows multi-frame voting outperforms single-frame prediction, with full fusion performing best by leveraging geometric and category priors to disambiguate challenging cases.

Finally, we analyze attachment refinement. Table~\ref{tab:ablation_all}(c) shows that relation attachment moves objects toward predicted support targets, room/wall repair reduces OOB, and final cleanup lowers collisions, yielding the highest overall physical plausibility.
\vspace{10pt}

\begin{center}
\small
\captionof{table}{Ablation results for the three pipeline modules.}
\label{tab:ablation_all}
\begin{minipage}[t]{0.32\textwidth}
\centering
\textbf{(a) View selection}\\[2pt]
\setlength{\tabcolsep}{12pt}
\resizebox{\linewidth}{!}{%
\begin{tabular}{l ccc}
\toprule
Variant & CD $\downarrow$ & F@5 $\uparrow$ & NC $\uparrow$ \\
\midrule
Largest Mask & 9.12 & \textbf{79.49} & 79.01 \\
3D Completeness & 8.45 & 68.46 & 76.91 \\
Full HierView & \textbf{8.40} & 68.99 & \textbf{79.90} \\
\bottomrule
\end{tabular}%
}
\end{minipage}
\hfill
\begin{minipage}[t]{0.32\textwidth}
\centering
\textbf{(b) Scene graph fusion}\\[2pt]
\setlength{\tabcolsep}{6pt}
\resizebox{\linewidth}{!}{%
\begin{tabular}{l ccc}
\toprule
Variant & Coll. Pairs $\downarrow$ & OOB Rate $\downarrow$ & Anchor $\uparrow$ \\
\midrule
Single Frame & 19.93 & 1.74\% & 0.494 \\
Multi-frame Vote & 17.29 & 1.70\% & 0.868 \\
Full Scene Graph & \textbf{16.10} & \textbf{0.90\%} & \textbf{0.892} \\
\bottomrule
\end{tabular}%
}
\end{minipage}
\hfill
\begin{minipage}[t]{0.32\textwidth}
\centering
\textbf{(c) Relation-guided attachment}\\[2pt]
\setlength{\tabcolsep}{12pt}
\resizebox{\linewidth}{!}{%
\begin{tabular}{l cc}
\toprule
Stage & Coll. Pairs $\downarrow$ & OOB Rate $\downarrow$ \\
\midrule
Relation Attach Only & 18.98 & 4.97\% \\
+ Room/Wall Repair & 19.55 & 0.90\% \\
Full & \textbf{16.10} & \textbf{0.90\%} \\
\bottomrule
\end{tabular}%
}
\end{minipage}
\end{center}
\FloatBarrier

\begin{table}[htbp]
  \centering
  \caption{Accuracy of Qwen-VL, GPT-4o, and Gemini-2.5-pro across different tasks and overall on the test set.}
  \label{tab:vqa-acc}
  \resizebox{\textwidth}{!}{%
    \begin{tabular}{lccccc}
      \toprule
      Model & Avg & Object Count & Distance Judgment & Relation Reasoning & Viewpoint Estimation \\
      \midrule
      \rowcolor{gray!15}
      \multicolumn{6}{l}{\textit{Open-source Models (Qwen series)}} \\
      Qwen2.5-VL-3B      & 0.5502 & 0.6279 & 0.5137          & 0.6130          & 0.4075 \\
      Qwen2.5-VL-3B-SFT  & \cellcolor{green!10}0.6309 & \cellcolor{green!10}0.6781 & 0.5000 & \cellcolor{green!10}0.6541 & \cellcolor{green!10}0.6678 \\
      \addlinespace
      Qwen2.5-VL-7B      & 0.6058 & 0.6644 & 0.5171          & 0.6678          & 0.5445 \\
      Qwen2.5-VL-7B-SFT  & \cellcolor{green!10}0.7648 & \cellcolor{green!10}0.8607 & \cellcolor{green!10}0.6507 & \cellcolor{green!10}0.8116 & \cellcolor{green!10}0.6884 \\
      \addlinespace
      Qwen2.5-VL-32B     & 0.5845 & 0.6210 & 0.5822          & 0.6164          & 0.5000 \\
      \addlinespace
      Qwen3-VL-2B        & 0.5479 & 0.5434 & 0.5377          & 0.6884          & 0.4247 \\
      Qwen3-VL-2B-SFT    & \cellcolor{green!10}0.6126 & \cellcolor{green!10}0.6644 & 0.4760 & 0.6610          & \cellcolor{green!10}0.6233 \\
     
      \addlinespace
      Qwen3-VL-8B        & 0.5997 & 0.6347 & 0.5890          & 0.6610          & 0.4966 \\
      Qwen3-VL-8B-SFT    & \cellcolor{green!10}0.6294 & \cellcolor{green!10}0.6918 & 0.4932 & \cellcolor{green!10}0.6918 & \cellcolor{green!10}0.6096 \\
      \addlinespace
      Qwen3-VL-32B       & 0.5822 & 0.6233 & 0.5548          & 0.6815          & 0.4486 \\
      \midrule
      \rowcolor{gray!15}
      \multicolumn{6}{l}{\textit{Closed-source Models}} \\
      GPT-4o              & 0.6345 & 0.6484 & 0.6096 & 0.7397 & 0.5342 \\
      Gemini-2.5-pro           & 0.9237 & 0.8995 & 0.9829 & 0.9760 & 0.8478 \\
      \bottomrule
    \end{tabular}%
  }
\end{table}

\subsection{Embodied VQA Evaluation}
\label{sec:embodied_vqa}
Beyond geometric fidelity and structural quality, we examine whether our
assembled scenes can serve as effective training environments for EAI. Using the reconstructed layouts
(\S\ref{sec:assembly}) and the scene-driven data generation pipeline
(\S\ref{sec:vqa}), we construct an embodied VQA dataset consisting
of 13 scenes and 5,688 question--answer pairs. Among them, 10 scenes
(4,374 samples) are used for training and 3 scenes (1,314 samples) for
evaluation, which cover object counting, distance judgment, relation
reasoning, and viewpoint estimation.

We fine-tune open-source Qwen-VL models on this dataset and compare them
against off-the-shelf closed-source models (Gemini-2.5-pro and GPT-4o) as shown in Table~\ref{tab:vqa-acc}. We find fine-tuning consistently improves performance across all open-source models. For example, Qwen2.5-VL-7B-SFT achieves 0.7648
average accuracy, a +15.9\% absolute gain over its base version. Similar gains are observed across other model
scales, indicating that the generated supervision is broadly effective. 

And the improvements are uneven across task types. Object counting and
viewpoint estimation benefit the most from supervision, while distance estimation remains relatively challenging, with only marginal
improvements or even slight degradation in some cases (e.g.,
Qwen2.5-VL-3B and Qwen3-VL-2B), suggesting that metric reasoning is
harder to capture from the generated data alone.

We also find fine-tuned open-source models become competitive with strong
closed-source baselines in several tasks. In particular,
Qwen2.5-VL-7B-SFT outperforms GPT-4o on object counting and relation
reasoning, and achieves comparable performance on viewpoint estimation.
However, Gemini-2.5-pro remains significantly stronger across all tasks,
especially on distance estimation and relation reasoning, highlighting
the remaining gap in general-purpose multimodal reasoning.

\section{Conclusion}
\label{sec:conclusion}
We presented ReScene, a framework for constructing simulation-ready indoor scenes from multi-view captures. By jointly modeling view selection, relation fusion, and physically grounded assembly, our method achieves strong performance in both reconstruction quality and structural plausibility, while enabling effective downstream embodied VQA. These results highlight the importance of structured, relation-aware scene representations for bridging real-world captures and simulation environments.

\bibliographystyle{unsrtnat}
\bibliography{references}

\appendix

\renewcommand{\thetable}{A\arabic{table}}
\setcounter{table}{0}
\renewcommand{\thefigure}{A\arabic{figure}}
\setcounter{figure}{0}

\section{Evaluation Protocol}

\subsection{Baseline Input Specifications}
\label{app:baseline_inputs}
We make all input differences across methods explicit:
\begin{enumerate}
    \item[(i)] all methods are given ground-truth camera parameters
    to ensure absolute-scale alignment;
    \item[(ii)] RICO additionally receives depth and normal
    supervision as required by its original setting;
    \item[(iii)] DRAWER is run on the full frame stream of each
    ScanNet sequence, since we observe that its reconstruction
    quality degrades substantially under subsampled inputs, and we
    therefore preserve its original dense-input regime;
    \item[(iv)] our method and the remaining multi-view baselines
    (SimRecon, RICO) use the same one-in-ten subsampled frames
    (typically 100--200 frames per scene);
    \item[(v)] the single-view methods Gen3DSR and SAM3D take a
    target image drawn from the same sequence.
\end{enumerate}
We note that DRAWER thus receives strictly denser observations than
our method, but we report its results faithfully rather than
restricting it to subsampled inputs that would disadvantage it
further.

\subsection{GPT-5 Perceptual Judge Protocol}
\label{app:gpt_judge}
To complement the low-level reconstruction and rendering metrics, we
use GPT-5 as a multimodal perceptual judge. For each scene, the model
is presented with rendered images from the reconstructed object-level
scene and prompted to score the results along three axes---visual
fidelity, scene completeness, and aesthetic quality---each on a 1--5
scale. The exact prompt template used for evaluation is provided
below.

\begin{verbatim}
You are an expert evaluator for object-level indoor 3D scene
reconstruction. You will be given one or more rendered images of a
reconstructed scene. Judge only the visible reconstruction quality in
the provided images.

Score the scene on three axes, each from 1 to 5:
1: very poor, 2: poor, 3: acceptable, 4: good, 5: excellent.

Axes:
- visual_fidelity: object geometry, texture/color plausibility,
  lighting consistency, and absence of obvious reconstruction artifacts.
- scene_completeness: whether the main furniture and room elements are
  present and recognizable, with few missing or duplicated objects.
- aesthetic_quality: overall visual coherence, natural arrangement,
  cleanliness of the layout, and plausibility as an indoor scene.

Ignore the image resolution and minor rendering noise. Penalize severe
floating objects, interpenetration, out-of-room objects, broken object
shapes, missing large furniture, and visually incoherent layouts.

Return only valid JSON in the following format:
{
  "visual_fidelity": {"score": <integer 1-5>, "comment": "<short reason>"},
  "scene_completeness": {"score": <integer 1-5>, "comment": "<short reason>"},
  "aesthetic_quality": {"score": <integer 1-5>, "comment": "<short reason>"}
}
\end{verbatim}

\section{Method Details}

\subsection{Relation Type Compilation}
\label{app:dispatch}

Table~\ref{tab:dispatch} specifies, for each supported relation type, the relation-specific energy $E_r(c, p)$ used in the relational fitting term of Eq.~\ref{eq:attachment}, together with the closed-form minimizer that defines $\mathrm{Compile}(r)(c, p)$. The last row covers the non-penetration term $E_{\mathrm{pen}}(i, j)$, which is applied as a final pass over non-related object pairs after relational fitting and topological propagation.

\begin{table}[t]
\centering
\caption{Compilation of relation types into energy terms and their closed-form minimizers.}
\label{tab:dispatch}
\small
\resizebox{\textwidth}{!}{%
\begin{tabular}{lll}
\toprule
\textbf{Relation type} & \textbf{Energy term $E_r(c, p)$} & \textbf{Closed-form minimizer} \\
\midrule
\texttt{rests-on} $\to$ floor & Signed distance from $c$'s lower contact & Snap $c$'s lower contact surface to the \\
& surface to the floor plane, plus the area & floor plane; project out-of-bound footprint \\
& of $c$'s footprint outside the floor polygon & back onto the floor polygon \\
\addlinespace
\texttt{mounts-on} $\to$ wall & Coplanarity residual + outward-normal & Resolve the canonical wall root to the wall \\
& mismatch between $c$'s back surface and & whose plane minimizes this energy; align $c$'s \\
& the selected wall, plus a lateral-order & back surface coplanar with that wall, \\
& violation penalty on each wall & preserving original lateral order \\
\addlinespace
\texttt{rests-on} $\to$ object & Signed distance from $c$'s contact surface & Project $c$'s contact surface onto a feasible \\
& to the nearest feasible support face of $p$ & support face of $p$ \\
\addlinespace
Any & $E_{\mathrm{pen}}(i, j)$: pairwise mesh & Apply small contact-reverse displacements \\
& interpenetration volume between & to non-related object pairs \\
& non-related objects $i, j$ & \\
\bottomrule
\end{tabular}
}
\end{table}
\subsection{Pipeline Variables and Data Flow}
\label{appendix:pipeline}

All stages of \methodname{} share a unified instance index $i$, taken from the multi-view reconstruction module~\citep{zust2025panst3rmultiviewconsistentpanoptic} as canonical IDs and either preserved or explicitly marked as missing by later stages. The reconstruction module produces, for each instance $i$, a set of 2D masks $\{m_{i,t}\}_{t=1}^{T}$ (where $m_{i,t}$ is a binary mask of instance $i$ in frame $t$, possibly empty when the instance is not visible) and a semantic point cloud $P_i \subset \mathbb{R}^3$ in the shared coordinate frame.

The downstream data flow is: HierView selects a single view index $t_i^\star \in \{1, \ldots, T\}$ per valid instance; SAM3D-Objects~\citep{sam3dobjects} consumes $(I_{t_i^\star}, m_{i,t_i^\star})$ and outputs the initial asset mesh $M_i^0$; registration aligns $M_i^0$ to $P_i$ to obtain the pose $T_i$; scene graph inference and attachment optimization share the same set of IDs.

Structural elements and low-evidence instances are excluded from asset generation, while \texttt{floor} and \texttt{wall} are represented as structural roots in the scene graph. Instances that fail asset generation or registration are explicitly marked as missing and are not used as attachable assets.

\paragraph{World-frame pose convention.} $\mathrm{Compose}(A, B) = A \cdot B$, which corresponds to first applying the parent's accumulated adjustment in the world frame and then applying the local update on top.

\subsection{HierView Thresholds}
\label{appendix:hierview}

The visibility stage of HierView admits a candidate view $v$ for instance $i$ only if its mask pixel count $\mathrm{area}(m_{i,v})$ is at least $a_{\min}$, and its bounding-box pixel count $|\mathrm{bbox}(m_{i,v})|$ relative to the total image pixel count $A_{\mathrm{img}}$ is at least $b_{\min}$. Both $a_{\min}$ and $b_{\min}$ are fixed across scenes. The semantic alignment threshold $\tau_s$ in Eq.~\ref{eq:sem} is also fixed.

Concrete values are $a_{\min}=100$ pixels, $b_{\min}=0.01$, and $\tau_s=0.12$. We keep candidates whose semantic rank is at most 3 and retain at least two candidates after semantic filtering when available. For completeness ranking, each instance contributes at most 30000 projected points.

\subsection{Registration Details}
\label{appendix:registration}

\paragraph{Two-stage alignment.} We initialize $T_i$ via FPFH feature matching~\citep{fpfh} with RANSAC, then refine with a Sim(3) ICP solver~\citep{umeyama}. Each iteration retains the closest fraction of nearest-neighbor correspondences (between sample points on $M_i^0$ and target points in $P_i$) and rejects pairs whose distance exceeds a small fraction of the target's spatial extent. The bounds in Eq.~\ref{eq:icp} are enforced during iteration: the per-step rotation update is clamped within $\theta_{\mathrm{step}}$, the accumulated refinement rotation is clamped within $\theta_{\mathrm{tot}}$ relative to $R_{i,0}$, and scale updates are clipped to $[s_{\min}, s_{\max}]$ after each step. We write the refined rotation as $R_i = \Delta R_i^{\mathrm{ref}} \, R_{i,0}$, where $R_{i,0}$ is the initial rotation produced by FPFH+RANSAC and $\Delta R_i^{\mathrm{ref}}$ is the accumulated refinement rotation.

\paragraph{Quality gating.} A refined alignment is accepted only if (i) it does not increase nearest-neighbor RMSE compared to the pre-refinement pose, (ii) the inlier ratio after refinement does not fall below the pre-refinement ratio by more than a small tolerance, and (iii) the final $s_i$, $R_i$, and $\Delta R_i^{\mathrm{ref}}$ lie within the bounds of Eq.~\ref{eq:icp}. Otherwise the pre-refinement pose is retained, catching any bound violation that the iterative clipping and clamping may have missed.

\paragraph{Failure fallback.} If the primary solver fails at the process level or fails to produce a valid output, we rerun the instance with a conservative nearest-neighbor ICP variant.

Concrete values are $\theta_{\mathrm{step}}=20^\circ$, $\theta_{\mathrm{tot}}=20^\circ$, $s_{\min}=0.90$, and $s_{\max}=1.10$. The initialization and final quality gates allow no RMSE worsening and at most a 0.01 inlier-ratio drop. ICP runs for at most 60 iterations, keeps correspondences below the 0.70 distance quantile after thresholding by 0.10 of the target extent, and requires at least 80 correspondences.

\subsection{Key-Frame Selection}
\label{appendix:keyframe}

Key frames for VLM inference are selected by greedy 3D coverage. We voxelize the instance point cloud at a fixed voxel size, project voxel centers into each candidate frame, and retain those whose projected pixel falls inside the 2D mask of the corresponding instance. We then greedily add the frame with the largest marginal gain in newly covered voxels until reaching a fixed frame budget.

Concrete values are a frame budget of 20 frames and a voxel size equal to the shortest side length of the sampled scene bounding box divided by 20, lower-bounded by $10^{-3}$m. The sampler uses at most 250000 global points with confidence at least 0.10. A frame is eligible when the mask-match ratio is at least 0.55 and at least five instances are visible; if fewer than 20 frames satisfy these filters, the sampler falls back to all frames with nonzero visible voxels.

\subsection{Multi-Frame Fusion Details}
\label{appendix:fusion}

\paragraph{Geometric prior $G_k$.} The geometric prior is selected by parent type:
\begin{itemize}
    \item \emph{Floor parent.} Floor contact: how close $c_k$'s lower contact surface is to the floor plane and how much of its footprint lies inside the floor polygon.
    \item \emph{Wall parent.} Wall alignment: minimum coplanarity residual and outward-normal agreement between $c_k$'s back surface and any candidate wall plane.
    \item \emph{Object parent.} Surface accessibility: whether $p_k$ exposes a feasible upward-facing support face large enough to host $c_k$'s contact footprint.
\end{itemize}
Each component is normalized to $[0, 1]$.

\paragraph{Category prior $\Phi_k$.} $\Phi_k$ is a normalized category-conditioned relation prior, derived from category co-occurrence statistics on a reference indoor-scene corpus. The raw co-occurrence count for the (child category, parent category, relation) triple is normalized into the unit interval, so $\Phi_k$ is neither a raw count nor a logit.

\paragraph{Fusion weights.} The weights $\alpha, \beta, \gamma$ in Eq.~\ref{eq:fusion} are fixed across all scenes.

\paragraph{A/B reranking.} When the top-2 score margin is small or the top score itself is low, we rerun a small number of high-visibility frames with an A/B-style prompt restricted to the top candidates, accepting the new result only if it agrees with the geometric prior.

Concrete values are $\alpha=0.60$, $\beta=0.25$, and $\gamma=0.15$. When a room-shell prior is available, we add it with weight 0.15 and rescale the vote/geometry/rule weights by 0.85. Edges are marked uncertain when the top-2 score margin is below 0.15 or the top score is below 0.65; the low-confidence reporting threshold is 0.60.
\section{Implementation Details}
\label{appendix:impl}

\subsection{End-to-End Pipeline}
\label{appendix:pipeline_impl}

ReScene is implemented as a staged pipeline. Given a scene ID and a directory of sampled RGB frames, the system runs the following steps:
\begin{enumerate}
    \item \textbf{Panoptic multi-view reconstruction.} We run PanSt3R to obtain camera-aware multi-view geometry, instance masks, per-instance labels, and semantic point clouds in a shared world coordinate frame.
    \item \textbf{HierView selection.} For each instance, we select one reconstruction view using the visibility, semantic, and 3D-completeness filters described in Sec.~\ref{sec:view_selection}.
    \item \textbf{Single-view asset generation.} The selected image and mask are passed to SAM3D-Objects to reconstruct an initial per-instance mesh.
    \item \textbf{Instance registration.} Each generated mesh is aligned to its corresponding semantic point cloud using bounded Sim(3) registration.
    \item \textbf{Scene graph construction.} We select key frames by greedy 3D coverage, render ID-overlaid images, infer per-frame relations with a VLM, and merge them into a global scene graph.
    \item \textbf{Attachment and refinement.} We compile relation edges into geometric constraints, run the attachment solver, optionally run a floor-only branch for dual attachment, and apply staged post-refinement and depenetration.
\end{enumerate}
All stages share the same instance IDs, so the selected view, generated mesh, registration target, scene graph node, and final attachment transform refer to the same object instance.

\subsection{HierView Details}
\label{appendix:hierview_impl}

HierView receives, for each instance $i$, all visible masks $\{m_{i,t}\}$, the instance label $\ell_i$, calibrated camera parameters, and the instance point cloud $P_i$. The implementation proceeds as:
\begin{enumerate}
    \item Build the candidate view set from frames where the instance has a non-empty mask.
    \item Remove candidates with mask area below $a_{\min}$ or bounding-box area ratio below $b_{\min}$.
    \item Encode the masked crop and the category text with CLIP, then keep views whose semantic score is above the threshold or whose semantic rank is sufficiently high.
    \item Project a capped set of instance points into each remaining view and compute the fraction landing inside the mask.
    \item Select the view with the highest 3D-to-2D completeness score. If a stage removes all candidates, the implementation falls back to the best candidate from the previous stage rather than dropping the instance immediately.
\end{enumerate}

\begin{table}[h]
\centering
\small
\caption{Default HierView hyperparameters.}
\label{tab:hierview_hparams}
\begin{tabular}{lc}
\toprule
Parameter & Value \\
\midrule
CLIP backend/model & ViT-B/32 \\
Semantic probability threshold $\tau_s$ & 0.12 \\
Semantic rank threshold & 3 \\
Semantic minimum keep count & 2 \\
Minimum mask area $a_{\min}$ & 100 pixels \\
Minimum bbox/image area ratio $b_{\min}$ & 0.01 \\
Maximum instance points for ranking & 30000 \\
\bottomrule
\end{tabular}
\end{table}

\subsection{Registration Details}
\label{appendix:registration_impl}

The registration stage aligns each generated object mesh $M_i^0$ to the semantic point cloud $P_i$ using a hybrid initialization-and-refinement strategy. Source points are sampled from the generated mesh and target points are taken from the PanSt3R instance point cloud. We first estimate a global candidate transform with FPFH features and RANSAC, then refine the accepted candidate with a bounded Sim(3) ICP solver. The ICP solver updates scale, rotation, and translation, but clips the scale and rotation update after each iteration.

The initialization candidate is accepted only if it does not worsen nearest-neighbor RMSE relative to identity and does not reduce the inlier ratio beyond the tolerance. The refined result is similarly gated against the pre-refinement pose. If the primary Open3D hybrid branch fails or produces an invalid result, we run a conservative nearest-neighbor ICP fallback.

\begin{table}[h]
\centering
\small
\caption{Default registration hyperparameters.}
\label{tab:registration_hparams}
\begin{tabular}{lc}
\toprule
Parameter & Value \\
\midrule
Similarity scale range & $[0.90, 1.10]$ \\
Scale regularization weight & 0.30 \\
Maximum per-iteration rotation update & $20^\circ$ \\
Maximum total residual rotation & $20^\circ$ \\
Initialization rotation gate & $50^\circ$ \\
Result rotation gate & $50^\circ$ \\
RMSE worsening tolerance & 0.00 \\
Inlier-ratio drop tolerance & 0.01 \\
Open3D voxel ratio & 0.02 of target extent \\
\bottomrule
\end{tabular}
\end{table}

\subsection{Key-Frame Selection and Scene Graph Prompting}
\label{appendix:keyframe_impl}

For scene graph inference, we greedily select up to 20 key frames by marginal 3D coverage. The pipeline voxelizes instance geometry, projects voxel centers into candidate frames, and counts newly covered voxels whose projections land inside the corresponding 2D instance masks. Each selected frame is rendered with visible instances overlaid by display IDs. The VLM is then prompted to output a strict JSON object:
\begin{verbatim}
{
  "objects": [
    {"id": <display_id>,
     "category": "<fixed_name>",
     "relation": "rests-on|mounts-on",
     "parent": <parent_id>}
  ]
}
\end{verbatim}
The allowed parents are visible object IDs plus two hidden structural roots for floor and wall. The category string is fixed by the input metadata and is not allowed to be changed by the model. The parser validates ID coverage, duplicate IDs, relation names, parent validity, self-parent loops, and graph cycles. If validation fails, error messages and the previous JSON are appended to a retry prompt. The default retry budget is 2, and the minimum accepted ID coverage is 0.95.

\subsection{Scene Graph Fusion}
\label{appendix:fusion_impl}

The fusion module first normalizes relation labels into two internal classes: support and attachment, corresponding to \texttt{rests-on} and \texttt{mounts-on} in the paper. For each instance, it aggregates category votes and candidate edge votes across frames. The default balanced score combines:
\begin{itemize}
    \item \textbf{Vote prior}: fraction of visible frames voting for a relation-parent pair.
    \item \textbf{2D geometry prior}: mask overlap and relative image position for candidate parent-child pairs.
    \item \textbf{3D geometry prior}: floor contact, wall proximity/alignment, or object support accessibility computed from instance statistics.
    \item \textbf{Category rule prior}: commonsense object-category preferences, e.g., beds and chairs usually rest on floor, pictures and switches usually mount on wall.
\end{itemize}
The default vote/geometry/rule weights are 0.60/0.25/0.15. Edges are marked uncertain when the top-2 score gap is below 0.15 or the top score is below 0.65. Uncertain edges are rechecked with an A/B prompt over the top two candidates; a changed answer is accepted only if it is not contradicted by the geometric margin.

\subsection{Attachment and Layout Refinement}
\label{appendix:attachment}

The attachment solver consumes the registered reconstruction JSON and the global scene graph. It converts each relation edge into a local geometric operation, then applies operations in topological order so that parent updates are propagated to descendants. The default refinement pipeline includes:
\begin{enumerate}
    \item \textbf{Floor snap}: snap floor-supported objects to the floor plane and project footprints back into the floor boundary if needed.
    \item \textbf{Wall resolution}: resolve the canonical wall root to a concrete wall plane by distance and orientation agreement.
    \item \textbf{Wall attachment}: align wall-mounted objects to the selected wall plane while preserving lateral order along the wall.
    \item \textbf{Object support}: project child contact surfaces onto feasible support faces of parent objects.
    \item \textbf{Near-wall snap}: optionally snap large furniture near walls when the scene graph and geometry support it.
    \item \textbf{Outside repair}: move objects whose footprints leave the room boundary back toward feasible interior positions.
    \item \textbf{Category-specific refinements}: apply conservative bed-wall forcing and cabinet-corner snapping when enabled.
    \item \textbf{Depenetration}: resolve residual mesh intersections by small displacements away from contact directions.
\end{enumerate}
Dual attachment runs both a wall-aware branch and a floor-only branch. This lets the system retain wall constraints when they are reliable while keeping a conservative fallback for objects whose wall relation is ambiguous.

\begin{table}[h]
\centering
\small
\caption{Default attachment hyperparameters.}
\label{tab:attachment_hparams}
\begin{tabular}{lc}
\toprule
Parameter & Value \\
\midrule
Default contact gap & 0.01m \\
Wall collision margin & 0.02m \\
Wall corner margin & 0.05m \\
Maximum floor-snap vertical translation & 1.40m \\
Maximum floor yaw rotation & $15^\circ$ \\
Maximum wall yaw rotation & $35^\circ$ \\
Near-wall auto-snap distance & 0.45m \\
Auto wall-snap maximum translation & 1.20m \\
Auto wall-snap maximum rotation & $25^\circ$ \\
Cabinet corner maximum translation & 0.20m \\
Depenetration iterations & 8 \\
Non-wall depenetration step & 0.04m \\
Wall-tangent depenetration step & 0.02m \\
\bottomrule
\end{tabular}
\end{table}

\section{Evaluation Details}
\label{appendix:evaluation}

\subsection{Geometry Metrics}
\label{appendix:geometry_metrics}

We evaluate geometry by uniformly downsampling reconstructed and reference point sets and computing bidirectional nearest-neighbor distances. Chamfer Distance is the average of prediction-to-reference and reference-to-prediction distances. Accuracy is the mean prediction-to-reference distance, while completion is the mean reference-to-prediction distance. Precision, recall, and F-score are computed at fixed distance thresholds, including 5cm and 10cm. Normal Consistency is computed from nearest-neighbor normal agreement when normals are available. Center offset measures the Euclidean distance between the reconstructed and reference scene centers.

\subsection{Structural Metrics}
\label{appendix:structural_metrics}

The structural metrics are computed from object meshes, final poses, and the fused scene graph together with scene-level floor/wall structural priors inferred by the pipeline. Collision count is the number of non-related object pairs with non-zero mesh intersection or voxelized overlap. Room containment pass rate is the fraction of object footprints inside the floor boundary after attachment. Floor contact pass rate measures whether floor-supported objects have bottom surfaces within a small tolerance of the floor plane. Wall contact pass rate measures whether wall-mounted objects are close and approximately parallel to the selected wall plane. Support contact pass rate measures whether object-supported children touch an accessible parent support surface. Attachment contact pass rate measures whether \texttt{mounts-on} edges satisfy the corresponding wall or parent contact condition. We report these metrics separately from geometry because a scene can have reasonable Chamfer Distance while still containing floating, penetrating, or out-of-room objects.

\subsection{VQA Data Generation}
\label{appendix:vqa_details}

The downstream VQA agent operates on the structured output of ReScene rather than on raw images. For each reconstructed scene, it parses object categories, centers, bounding boxes, support edges, attachment edges, and structural coordinates. It then generates question-answer pairs in five categories: object recognition, relative position, support relation, attachment relation, and multi-hop spatial reasoning. Questions are generated programmatically from scene graph templates and filtered to avoid ambiguous answers, such as pairs with nearly identical distances or objects whose relation confidence is below a threshold. This design ensures that the generated data tests the explicit structure recovered by ReScene.

\section{Limitations}
\label{appendix:limitations}

ReScene depends on the quality of the upstream multi-view reconstruction, instance masks, and semantic point clouds. When objects are missed, severely over-segmented, or consistently occluded in the input, later stages can mark the instance as missing or reject unreliable assets, but cannot fully recover object structure that is absent from the upstream evidence.

The current scene graph focuses on support and attachment relations, mainly \texttt{rests-on} and \texttt{mounts-on}. This relation vocabulary covers the dominant physical constraints in many indoor scenes, but does not yet model finer semantic relations such as containment, part-whole structure, articulation, or functional affordances.

The attachment solver assumes a coarse room shell with dominant floor and wall planes and uses rule-based relation compilation. This design improves stability and interpretability, but performance can decrease in overly complex rooms, such as scenes with highly irregular geometry, non-planar or curved walls, multi-level layouts, dense clutter, or unusual furniture arrangements that violate the assumed support and wall-attachment patterns.

Finally, our evaluation is conducted on a finite ScanNet subset. Although the selected scenes cover diverse indoor layouts, broader validation on larger-scale captures, more complex room types, and noisier casual videos remains an important direction for future work.

\end{document}